\documentclass[sigconf, nonacm]{acmart}
\renewcommand\footnotetextcopyrightpermission[1]{} 

\usepackage[utf8]{inputenc} 
\usepackage[T1]{fontenc}    
\usepackage{hyperref}       
\usepackage{url}            
\usepackage{booktabs}       
\usepackage{amsfonts}       
\usepackage{nicefrac}       
\usepackage{microtype}      
\usepackage{xcolor}         
\usepackage{verbatim}
\usepackage{multirow}
\usepackage{graphicx}
\usepackage{caption}
\usepackage{subcaption}
\usepackage{array}
\newcolumntype{?}{!{\vrule width 1pt}}
\usepackage{pifont}
\usepackage{amsmath}
\usepackage{algorithm,algpseudocode}
\usepackage{graphicx}
\usepackage[normalem]{ulem}
\useunder{\uline}{\ul}{}

\AtBeginDocument{%
  \providecommand\BibTeX{{%
    \normalfont B\kern-0.5em{\scshape i\kern-0.25em b}\kern-0.8em\TeX}}}

\DeclareUnicodeCharacter{2212}{-}
\begin{document}

\pagestyle{plain} 
\title{Shapley-Value-Based Graph Sparsification for GNN Inference}
\author{Selahattin Akkas}
\orcid{0000-0001-8121-9300}
\affiliation{%
  \department{Department of Intelligent Systems Engineering}
  \institution{Indiana University Bloomington}
  \city{Bloomington}
  \state{Indiana}
  \country{USA}
}
\email{sakkas@iu.edu}

\author{Ariful Azad}
\orcid{0000-0003-1332-8630}
\affiliation{%
  \department{Department of Computer Science \& Engineering}
  \institution{Texas A\&M University}
  \city{College Station}
  \state{Texas}
  \country{USA}
}
\email{ariful@tamu.edu}

\renewcommand{\shortauthors}{Akkas and Azad}

\begin{abstract}
Graph sparsification is a key technique for improving inference efficiency in Graph Neural Networks by removing edges with minimal impact on predictions. GNN explainability methods generate local importance scores, which can be aggregated into global scores for graph sparsification. However, many explainability methods produce only non-negative scores, limiting their applicability for sparsification. In contrast, Shapley value based methods assign both positive and negative contributions to node predictions, offering a theoretically robust and fair allocation of importance by evaluating many subsets of graphs. Unlike gradient-based or perturbation-based explainers, Shapley values enable better pruning strategies that preserve influential edges while removing misleading or adversarial connections. Our approach shows that Shapley value-based graph sparsification maintains predictive performance while significantly reducing graph complexity, enhancing both interpretability and efficiency in GNN inference.
\end{abstract}

\maketitle

\let\thefootnote\relax\footnotetext{Accepted at the KDD 2025 Workshop: Machine Learning on Graphs in the Era of Generative Artificial Intelligence (MLoG-GenAI).}

\section{Introduction}
Graph Neural Networks (GNNs) have become very popular in the field of machine learning, specifically in handling graph-structured data \cite{wu2020comprehensive}. Unlike traditional neural networks, GNNs leverage the hidden relationships within graph data, making them one of the preferred methods for social networks \cite{min2021social, li2023survey-social}, recommendation systems \cite{gao2022graph-recommend, wang2024distributionally-recommend}, molecular modeling \cite{wu2020comprehensive, khemani2024review}, and financial fraud detection \cite{motie2024financial, cheng2025financialfraud}. GNNs' ability to capture local and global patterns through message passing has significantly improved predictive model performance in these domains \cite{zhou2020graph}.

Since the weight matrices in GNNs are typically small, the computational and memory complexity of GNN inference is often dominated by the size of the graph, that is, the number of nodes and edges. As real-world graphs continue to grow in size and structural complexity, the scalability and feasibility of GNN inference on memory-constrained edge devices and GPUs become increasingly challenging.
To address this issue, various graph sparsification techniques \cite{ugs_chen21, earlybird_you22, wdglt-2023, inductlottery-sui2024, fastglt-yue2024} have been introduced in the literature. 
Most of these methods are inspired by the Lottery Ticket Hypothesis, which suggests that a subnetwork of a GNN consisting of a subset of parameters, layers, nodes, and/or edges can be trained to achieve performance comparable to that of the full model. These approaches, therefore, focus on identifying and pruning redundant edges to reduce graph complexity, lower resource consumption, and accelerate inference.

A second class of methods also seeks to identify uninformative nodes and edges, but with the goal of explaining the predictions made by GNN models~\cite{ying2019gnnexplainer, luo2020pgexplainer, vu2020pgmexplainer, duval2021graphsvx}. 
The GNN explanation methods aim to identify crucial subgraphs contributing more to the predictions. The fidelity~\cite{2023yuantaxonomy} metric is commonly used to evaluate the success of GNN explanation methods. It measures how model predictions change when some edges are removed. Specifically, $Fidelity_{+}$ measures how the model prediction changes when important edges are removed, while $Fidelity_{-}$ measures how the model prediction changes when the least important edges are removed. The ability to identify the most and least important edges for GNN explanation methods motivated us to apply explanation scores to graph sparsification. Graph sparsification using explanation scores offers several advantages: (1) it eliminates the need to retrain the model after sparsification, (2) it avoids the need to repeatedly recompute edge scores to determine the optimal sparsity level, and (3) the inference is inherently explainable as the graph is already sparsified but removing unimportant edges.

While the fidelity score is a good metric for evaluating GNN explanation methods, it has limitations. The fidelity metric only focuses on the magnitude of score change in the model prediction; it does not consider whether the model prediction improves or worsens. Many GNN explanation methods generate only non-negative explanation scores to define importance. However, some edges significantly reduce model prediction, but those edges are considered important since they significantly change the fidelity score. Explanation methods that only give non-negative scores have this shortcoming. We argue that an explanation score that can provide positive and negative importance scores should perform similarly or better on graph sparsification tasks.

\begin{figure*}[!t]
    \centering
        \includegraphics[width=0.40\linewidth]{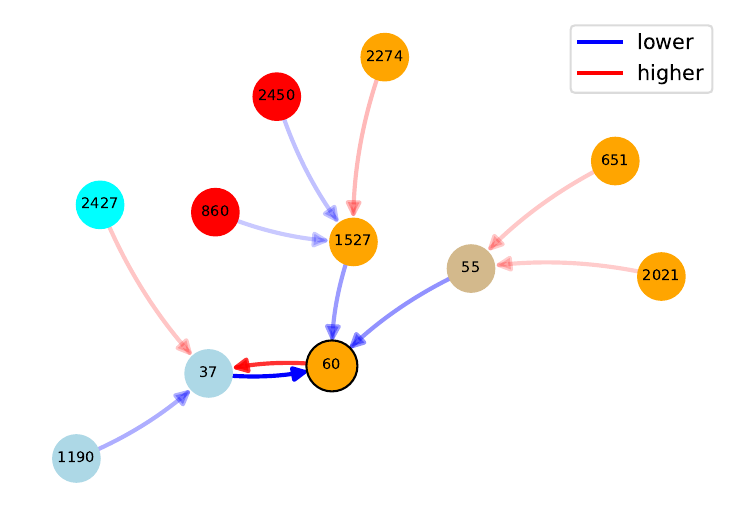}\hfill
        \includegraphics[width=0.46\linewidth]{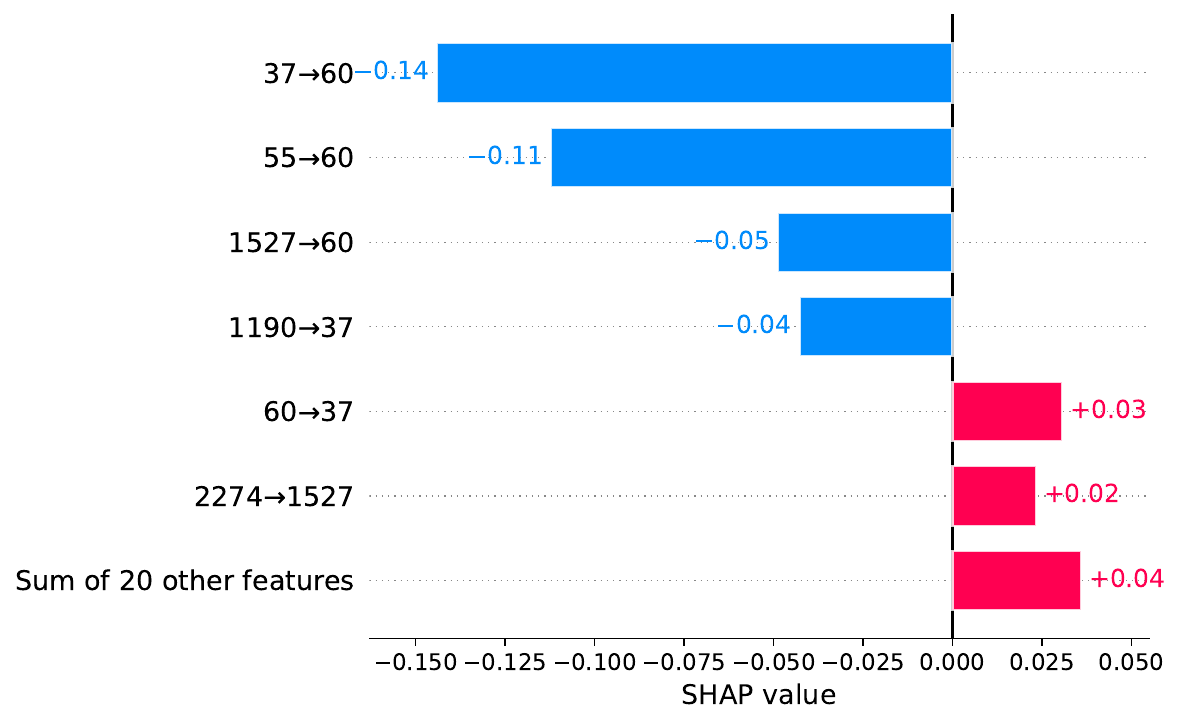}
        \caption{Example Shapley value explanation on Cora node 60. Node colors denote classes. The bar chart on the right shows important edges and their Shapley values. While red colors show a positive contribution, blue colors show a negative contribution.}
        \label{fig:sparse-intro}
\end{figure*}

Shapley value based GNN explanation methods~\cite{yuan2021subgraphx, mastropietro2022edgeshaper, perotti2023graphshap, akkas2024gnnshap} provide both positive and negative attribution scores in addition to high-quality explanations. Figure \ref{fig:sparse-intro} shows an example of Shapley values for a node. In the figure, the edge from node 37 to node 60 is the most important edge; however, removing this edge improves model performance. The ability to prune the graph's negatively attributing and least significant edges without compromising the accuracy allows for higher sparser graphs and faster inference.

In this paper, we explore graph sparsification using Shapley values and demonstrate its unique advantages over existing explanation methods for node classification tasks, particularly in reducing the computational complexity of inference.
Overall, the paper makes the following contributions:
\begin{itemize}
    \item We comprehensively evaluate Shapley value explanations across multiple datasets and models, demonstrating its robustness and generalizability on graph sparsification.
    \item We compare state-of-the-art GNN explanation methods and sparsification techniques, highlighting superior performance in maintaining model accuracy.
    \item We also compare our sparsification approach with graph lottery ticket approaches, demonstrating competitive or improved sparsification ratios while maintaining model performance.
    
\end{itemize}

\section{Background \& Related Work}
We denote a graph as $G = \left \{ V, E \right \}$, where $V$ is the set of $N$ nodes, $E$ is the set of edges, and $X {\in} \mathbb{R}^{N \times F} $ is the node feature matrix.  
$A{\in}\{0,1\}^{N\times N}$ is the binary adjacency of the graph, where $A_{ij} = 1$ if ${v_i, v_j} \in E$, and $A_{ij} = 0$ otherwise.
Let $y=\{y_1, y_2, ..., y_N\}$ denote the labels of the nodes, where each label $y_i$ belongs to one of 
$\mathbb{C}$ classes in a multiclass node classification task. An $l$-layered GNN model $f$ takes $X$ and $A$ as input and generates predictions for the $i$th node:  $\hat{y_i}=f(X, A)$, where $\hat{y_i} {\in} \mathbb{C}$.

{\bf Computational graph.} When predicting the class of a node, the GNN inference needs a small subgraph of the entire graph. 
Specifically, the prediction of a node $v$ with an  $ l$-layer GNN only depends on $v$'s 1-hop through $l$-hop neighbors, the edges among them, and any associated node and edge features. 
This $l$-hop subgraph is referred to as the computational graph $G_c(v)$, which contains all the necessary information for predicting $v$.

\subsection{GNN Explanations}
A GNN explanation method $\Phi$ generates explanations for a given node $v$ with respect to a target class $t \in \mathbb{C}$. The target class may correspond to either the ground truth label or the predicted class. 
Popular GNN explanation methods aim to explain the prediction for node $v$ by taking as input a trained GNN model $f$ and the node’s computational graph $G_c(v)$. The explanation typically consists of a small subgraph $G_s(v) \subseteq G_c(v)$ and/or a subset of node features that most significantly influence the prediction. The key idea is to retain only the edges and features that contribute most to the model’s decision.
In this work, we focus exclusively on subgraph-based explanations, as our goal is to sparsify graphs by pruning edges. The explanation model assigns an importance score $\phi_{v}^t(i,j)$ to each edge $(v_i, v_j)$, indicating its contribution to node $v$’s prediction for the target class $t$. Depending on the explanation method, these scores can be either positive or negative.

\subsection{Related Work}
Graph sparsification aims to remove edges from a graph while preserving the model's predictive performance. Various approaches have been proposed in the literature to achieve this goal.

Denoising methods, such as NeuralSparse \cite{neuralsparse-zheng20d} and PTDNet \cite{ptdnet-luo21}, aim to enhance the GNN's generalization capability by reducing noise and making the GNN less sensitive to the graph's quality. NeuralSparse learns irrelevant edges during training and removes them to improve node representations, whereas PTDNet utilizes a probabilistic edge dropout mask to learn noisy edges and subsequently drops them.

Graph lottery ticket approaches aim to find a sparser graph and model parameters with similar or better accuracy than the original graph and model, which are called winning tickets. UGS \cite{ugs_chen21}, Early-Bird GCNs \cite{earlybird_you22}, WD-GLT \cite{wdglt-2023}, CGP \cite{comprehensivegradual-liu2024},  ICGP \cite{inductlottery-sui2024}, FastGLT \cite{fastglt-yue2024} and \cite{lotteryautomated-zhang24, mog-zhang2025, twohead-zhang24, trainprune-ma2024} aim to find the winning tickets. While these approaches can sparsify graph and model parameters, shallow GNN models (e.g., 2-4 layers) usually work well. Moreover, most of these approaches require retraining for each target sparsity level. Since models are shallow, starting with a small model and experimenting with larger or deeper models is more practical. In addition, they only use a small portion of nodes (the training set) in semi-supervised GNN learning, which has a limited impact on graph sparsification.

Explainability based graph sparsification approaches use explanation scores to sparsify the graph. EEGL \cite{eegl-naik24} applies frequent subgraph mining to find the most common patterns using GNNExplainer. Then, patterns are used as additional features. The authors train the model iteratively with found subgraphs to improve the model's accuracy. While EEGL finds subgraphs, it mainly focuses on enhancing the model's performance rather than sparsification.

IGS \cite{igs-li2023} focuses on brain graph sparsification. It iteratively learns a trainable edge mask during training and removes unimportant edges. xAI-Drop \cite{xaidrop-luca24} computes node explanations and drops nodes based on the explainability score. It first considers the model's probability for candidate nodes. Then, it computes explainability scores. Finally, it applies Bernoulli-based node drops. While these works utilize explainability, their primary target is to enhance model accuracy during training. We focus on enhancing the inference speed of the pre-trained model.

The work by Shin et al.~\cite{fidprune-shin24} is closely related to ours, as it leverages GNN explanations for graph sparsification.  
The authors propose a fidelity-inspired pruning method, where the 
$Fidelity_{-}$score measures the change in model prediction when non-essential edges are removed. 
They aggregate individual edge explanation scores into global importance scores and prune edges with the lowest values. However, their method considers only non-negative edge scores. 
While $Fidelity_{-}$  effectively identifies unimportant edges, we show that a better sparsification can be achieved by removing edges with negative importance (i.e., edges that reduce prediction confidence).

\section{Method}
\begin{figure*}[!t]
  \centering
  \includegraphics[width=1.0\textwidth]
  {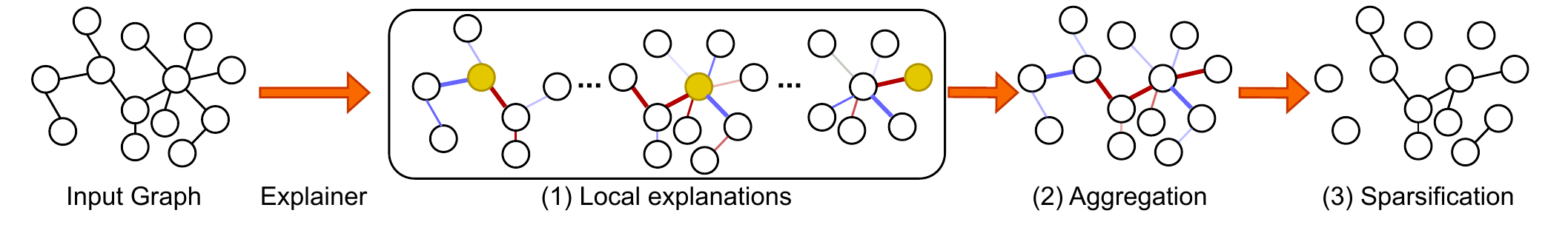}
    \caption{Overview of the explanation based graph sparsification algorithm. Firstly, explanation scores are computed for each node. Secondly, the scores are aggregated for each edge. Finally, using the aggregated scores, edges are pruned until the target sparsity ratio is reached.}
  \label{fig:sparse-algo}
\end{figure*}

\subsection{Shapley Values}

Shapley value \cite{shapley1951notes} is a game-theoretic method that fairly distributes gains to collaborating players. A Shapley value GNN explanation method considers nodes or edges as \textbf{players} and fairly distributes the model output to players.

The exact Shapley value of a player is computed by Eq.~\ref{eq:sparse-shapley-exact}, where $n$ denotes the number of players, S a coalition (a subset of players), and $f(S\cup \{i\}) - f(S)$ is player $i$'s marginal contribution to coalition $S$. Shapley values can be positive and negative; while positive values increase model prediction probability, negative values decrease.

\begin{equation}
\phi_{i}= \sum_{S\subseteq \{1,...,n\} \setminus \{i\}}^{2^{n-1}}\frac{|S|!(n - |S| -1)!}{n!}\left [f(S\cup \{i\}) - f(S)  \right ]
\label{eq:sparse-shapley-exact}
\end{equation}

Computing exact Shapley values is impractical when the number of players is large, as it requires evaluating $2^{n}$ coalitions. \cite{duval2021graphsvx, akkas2024gnnshap} use a simple surrogate model to compute the approximation of Shapley values using a much smaller subset of coalitions ($k \ll 2^{n}$). The surrogate model $g$ is defined in Eq. \ref{eq:sparse-surrogate-model}, where $m \in \{0, 1\}^{1xn}$ denotes a binary coalition mask, $S$, and $\phi$ are model parameters: the approximation of the Shapley values. 

\begin{equation}
f(x) \approx g(x) = \phi_0 + \sum_{i=1}^{n} \phi_i m_{i},
\label{eq:sparse-surrogate-model}
\end{equation}

\subsection{Graph Sparsification by Explanation Scores}
Most GNN explanation methods give local explanations, i.e, an explanation for a node's classification. Since the same edges are in many different nodes' $l-hop$ neighborhoods, there are multiple scores for each edge. To get a global score for each edge, we need to aggregate scores. In this work, we use \textbf{mean} aggregation, where we calculate the average score for each edge. We show our sparsification algorithm in Algorithm \ref{alg:sparse} and Figure \ref{fig:sparse-algo}.

We also considered the sum and weighted mean aggregations. We use model predictions as probabilities in the weighted mean, thinking they should have less weight when the model is less sure. However, we don't see significant differences in pruning performance. The sum and weighted mean results are provided in Appendix \ref{sec:sparse-appendix-more-aggr}.

\begin{algorithm}[!t]
\caption{GNN Explanation-Based Graph Sparsification}
\label{alg:sparse}
\begin{algorithmic}[1]
\Require Graph $G = (V, E)$, edge importance scores $S_v(e)$ for each node $v \in V$ and edge $e \in E$, sparsification threshold $\tau$
\Ensure Sparsified graph $G' = (V, E')$
\State Initialize empty set $E' \gets \emptyset$
\State Initialize edge score map $S \gets \emptyset$
\For{each edge $e \in E$}
    \State $S(e) \gets \frac{1}{|V_e|} \sum_{v \in V_e} S_v(e)$
    \Comment{$V_e$: nodes utilize edge $e$}
\EndFor
\State Sort edges $e \in E$ by $S(e)$ in descending order into list $L$
\For{each edge $e$ in $L$}
    \If{$|E'| < (1 - \tau) \cdot |E|$}
        \State $E' \gets E' \cup \{e\}$
    \EndIf
\EndFor
\State \Return $G' = (V, E')$
\end{algorithmic}
\end{algorithm}

\section{Experiments}
We hypothesize that Shapley-based explanation methods are particularly effective for graph sparsification.
While several Shapley-based GNN explanation methods have been proposed, we use GNNShap~\cite{akkas2024gnnshap}, a recent method that has demonstrated superior performance compared to other approaches.
In our experiments, we compare GNNShap-based sparsification with other explanation methods. 
We also compare Shapley-based sparsification with graph lottery ticket (GLT) baselines. 
In each experiment, we apply different sparsification methods to the graph, perform GNN inference on the test nodes, and report the resulting test accuracy. 
A sparsification method is considered more effective if it maintains high test accuracy despite sparsification of the graph.
For GNN explanations, we compute explanations for each node for predicted classes and aggregate explainability scores. We repeat each experiment five times and provide the average results.

\subsection{Datasets}
In our experiments, we utilize three well-known real-world citation datasets: Cora, CiteSeer, and PubMed \cite{dataset_planetoid}, as well as a coauthorship dataset: Coauthor-CS \cite{dataset_coauthor}. In citation datasets, nodes represent papers, and edges represent citations. Node features are bag-of-words vectors, the most common words in the documents. Coauthor-CS is a coauthorship graph where nodes are authors and edges show coauthorships. Its node features are keywords in the papers. Since there is no public train, validation, and test split for Coauthor-CS, we randomly sample 30 nodes from each for the training and validation set, while using the remaining for testing. Table \ref{table:sparse-dataset-summary} shows summaries of datasets.

\begin{table}[!t]
\centering
\caption{Dataset Summaries}
\label{table:sparse-dataset-summary}
\begin{tabular}{lcccc}
\hline
Dataset          & Nodes & Edges  & Features & Classes\\
\hline
Cora           & 2708  & 10556  & 1433     & 7\\
CiteSeer            & 3327  & 9104  & 3703     & 6\\
PubMed         & 19717 & 88648  & 500      & 3\\
Coauthor-CS         & 18333 & 163788  & 6805  &15\\
\hline
\end{tabular}
\end{table}

\begin{table}[!t]
\centering
\caption{Trained GNN models and their training, validation, and test set accuracies.}
\label{table:sparse-model-acc}
\begin{tabular}{llccc}
\hline
Model                & Dataset  & Train  & Validation   & Test  \\
\hline
\multirow{3}{*}{GCN} & Cora     & 100.00 & 79.40 & 81.50 \\
                     & CiteSeer & 99.17  & 70.40 & 71.00 \\
                     & PubMed   & 100.00 & 80.20 & 78.80 \\
                     & Coauthor-CS   & 95.33 & 93.33 & 91.99 \\
\hline
\multirow{3}{*}{GAT} & Cora     & 100.00 & 79.40 & 81.40 \\
                     & CiteSeer & 99.17  & 73.20 & 71.60 \\
                     & PubMed   & 98.33  & 80.60 & 78.50 \\
                     & Coauthor-CS   & 94.22 & 92.22 & 91.29 \\
\hline
\end{tabular}
\end{table}

\subsection{Models}
We use a two-layer GCN \cite{kipf2016semi_GCN} and GAT \cite{velickovic2017graph_gat} models with 16 hidden layer sizes for Cora, CiteSeer, and PubMed, and 64 for Coauthor-CS datasets. In the GAT models, we use eight attention heads. We use 0.5 dropout and ReLU as the activation function. We train the model for 200 epochs using a learning rate of 0.01. For GLT experiments, we follow UGS, using the same parameters with a 512 hidden layer size. Table \ref{table:sparse-model-acc} shows model training, validation, and test accuracies.

\subsection{Baselines}
\subsubsection{GNN explanation methods}
\begin{itemize}
    \item Saliency \cite{pope2019explainability, baldassarre2019explainability}: uses the absolute values of gradients with respect to edges as scores.
    \item Guided Backpropagation \cite{guidedbackprop-springenberg15}: also uses gradients, except negative gradients are pruned in the backpropagation.
    \item Integrated Gradients \cite{sundararajan2017integratedgrads} computes the gradients of the model's output with respect to edges, tracing a path from a baseline to the actual input.
    \item GNNExplainer \cite{ying2019gnnexplainer}: uses mutual information to learn edge scores. It uses a learnable mask and trains it iteratively using gradients to maximize the mutual information.
    \item PGExplainer \cite{luo2020pgexplainer}: also utilizes mutual information.  It trains a neural network model to generate edge scores.
    \item FastDnX \cite{pereira2023fastdnx}: utilizes linear surrogate model based on the SGC \cite{SGC_wu2019} to explain GNN models.
    \item GraphSVX \cite{duval2021graphsvx} is a Shapley value based GNN explanation method that uses a linear surrogate model to approximate Shapley values.
    \item GNNShap \cite{akkas2024gnnshap} is another Shapley value method specifically designed for GNNs. While it is similar to GraphSVX, it generates explanation scores for edges and utilizes a GPU for coalition sampling and model predictions. Therefore, it is an order of magnitude faster than GraphSVX and can evaluate more coalition samples.
    
\end{itemize}

FastDnX and GraphSVX generate scores for nodes. We convert node scores to edges by averaging the scores of connected edges.

While there are other Shapley value based GNN explanation methods, GraphShap~\cite{perotti2023graphshap} and EdgeSHAPEr~\cite{mastropietro2022edgeshaper} are designed for graph classification; SubGraphX~\cite{yuan2021subgraphx} can only be used for small graphs. Therefore, we utilize GraphSVX and GNNShap in our work as two representative Shapley value-based GNN explanation methods.

\subsubsection{Graph lottery ticket baselines}
\begin{itemize}
    \item Unified GNN Sparsification (UGS) \cite{ugs_chen21}: iteratively prunes the GNN model and the adjacency matrix to find a smaller model and graph that gives similar or higher accuracy. Then, it trains the pruned model with the pruned adjacency matrix. We disable model pruning and use the same parameters provided in the source code for UGS.
    \item WD-GLT \cite{wdglt-2023}: addresses the limitation of UGS, which only considers a fraction of the adjacency matrix in the loss. WD-GLT adds the Wasserstein Distance (WD) \cite{villani2009wasserstein} between nodes predicted to be in the same class to the loss.
    \item FastGLT \cite{fastglt-yue2024}: proposes a one-shot pruning method as a faster alternative to iterative pruning, achieving higher sparsity and faster speeds. It starts with a pre-trained model and removes model weights and edges in a single step based on their magnitude. Finally, a denoising process is applied to minimize the effect of noise introduced during pruning.
\end{itemize}

\section{Results}
\subsection{Experimental Results with Explanation Methods}

\begin{figure*}[!t]
  \centering
\includegraphics[width=0.98\textwidth]{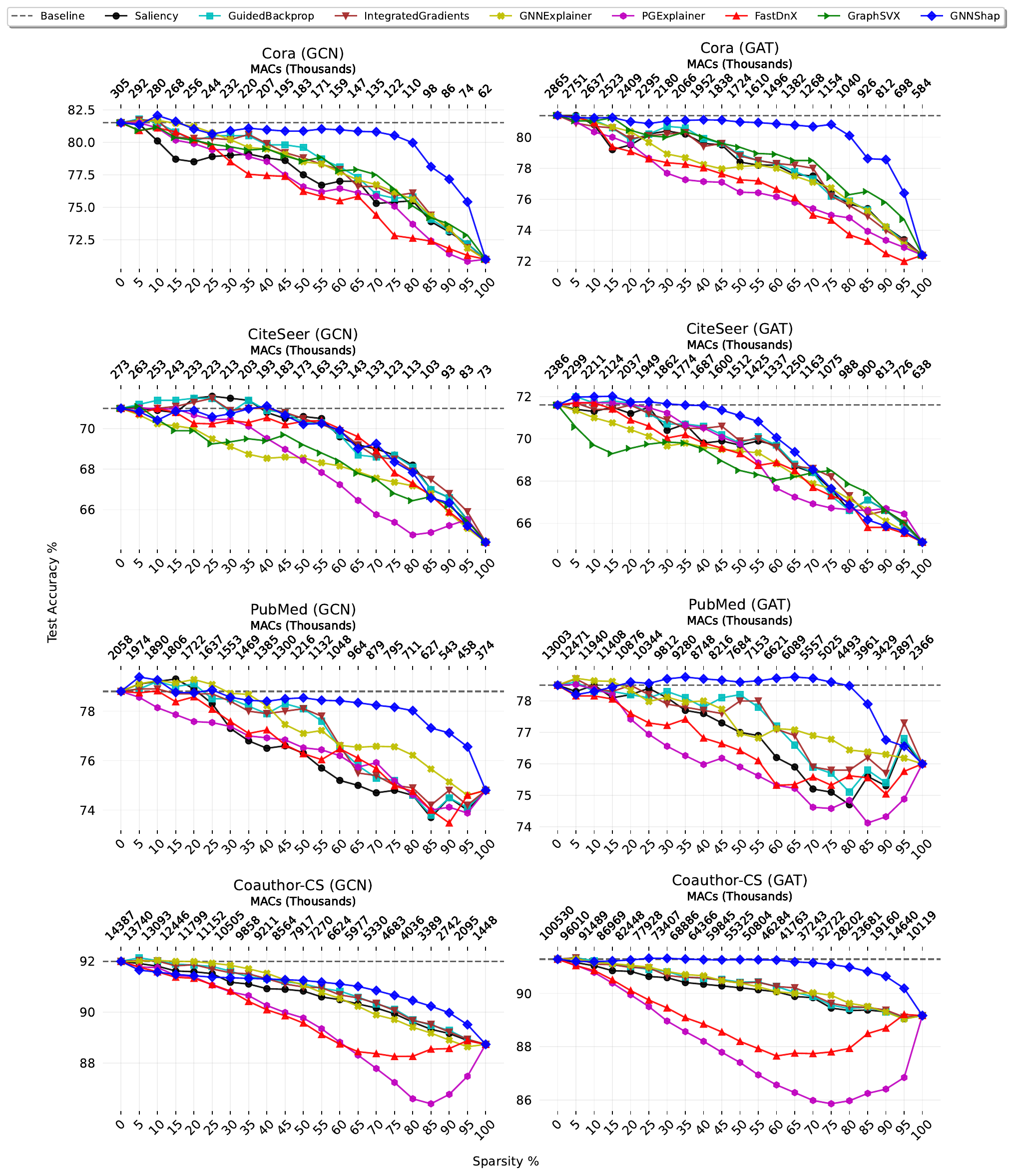}
    \caption{Test accuracies when edges sparsified using \textbf{mean} aggregated explanation scores. GNNShap gives competitive or even better accuracies for high sparsification percentages.}
  \label{fig:sparse-meanres}
\end{figure*}
We present our graph sparsification results using various explanation methods in Figure~\ref{fig:sparse-meanres}. 
The figure demonstrates that the Shapley value-based method, GNNShap, consistently achieves higher test accuracy at high sparsification levels. 
For example, on the Cora dataset using both GCN and GAT models, GNNShap can prune 80\% of the edges with less than a 2\% drop in accuracy.
Similarly, on the PubMed dataset with the GCN model, GNNShap can prune 80\% of the edges with less than a 2\% drop in accuracy.
\begin{figure*}[!t]
  \centering
    \includegraphics[width=1.0\textwidth]{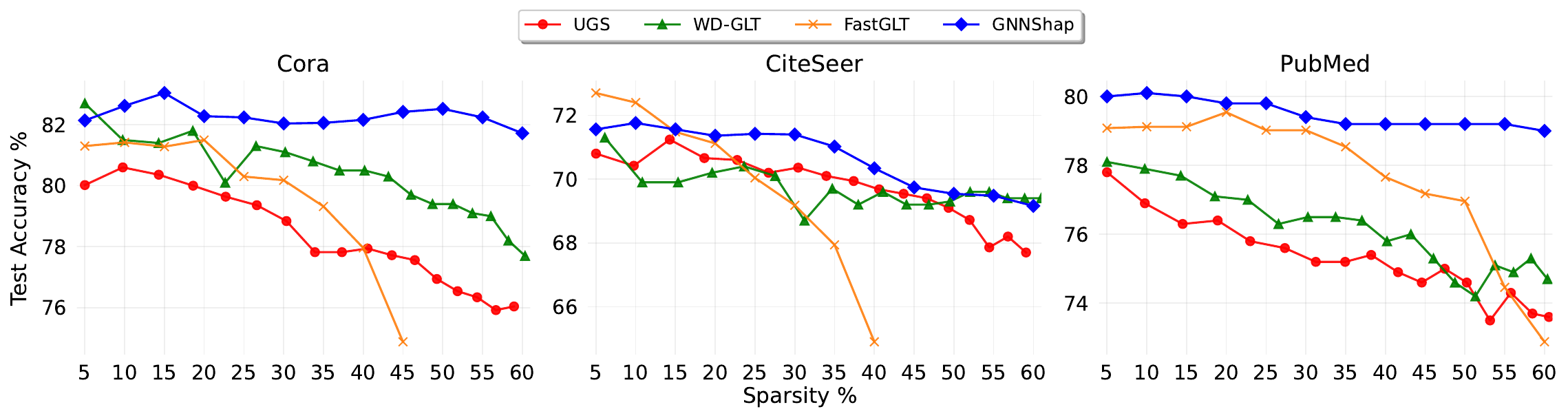}
    \caption{Test accuracy comparison of two-layer GCN model with 512 hidden layer size. Shapley value-based sparsification achieves higher accuracy with significantly less loss in accuracy.}
  \label{fig:sparse-gltcompare}
\end{figure*}
Notably, on PubMed and Coauthor-CS with the GAT model, GNNShap matches the original accuracy even after pruning 80\% and 55\% of the edges, respectively.

Overall, GNNShap enables significantly higher pruning rates with minimal accuracy loss compared to other explanation methods.
One notable exception is the CiteSeer dataset with the GCN model, where all explainers perform similarly. 
This is likely due to the lower test accuracy of the model on CiteSeer. 
GNNShap’s superior performance can be attributed to (i) its ability to better distinguish between important and unimportant edges, and (ii) its capacity to assign both positive and negative attribution scores to edges.

To evaluate the computational efficiency of Shapley value-based sparsification, we report the number of Multiply-Accumulate operations (MACs) required during the message-passing step of GNN inferences. For example, on the Cora dataset using a GCN model, inference on the original graph requires 305,000 MACs. With 80\% edge pruning, this is reduced to 110,000 MACs, a 64\% reduction in message-passing computation.
Similarly, for the GAT model on Cora, the baseline requires 2,865,000 MACs, which drops to 1,040,000 MACs (a 64\% reduction) at 80\% 
On PubMed with GCN, GNNShap reduces MACs from 2,058,000 to 711,000 (at 80\% sparsity). 
For PubMed with GAT, the baseline requires 13,003 MACs, which is reduced to 4,493 MACs at 80\% sparsity (a 65\% reduction in computation). On Coauthor-CS with GAT, GNNShap reduces MACs from 100,530 to 50,804 at 55\% sparsity, resulting in a 49\% reduction. These significant reductions in MACs highlight the ability of Shapley value-based sparsification to maintain high accuracy while substantially lowering the computational cost of message passing.

While there are no significant differences among the other explainers, overall, PGExplainer tends to perform the worst. GrapSVX is also based on Shapley value and generally provides high-quality explanations; however, its consideration of nodes as players requires score conversion by averaging the scores of two connected nodes, which limits its sparsification performance. 
Moreover, we do not have GraphSVX results for the PubMed and Coauthor-CS datasets, as GraphSVX was unable to generate all node explanations within the 10-hour time limit.

\subsection{Experimental Results with GLT Methods}
In this section, we compare Shapley value-based sparsification (GNNShap) with GLT methods. Figure \ref{fig:sparse-gltcompare} shows test accuracies of GLT methods and GNNShap. UGS only utilizes training nodes in the gradient computation and learning process. Since training nodes are a small subset of the data, UGS has gradient information on a limited number of edges. For instance, 140 out of 2708 nodes were used in the training on the Cora dataset. Therefore, its graph pruning performance will be limited as shown in Figure \ref{fig:sparse-gltcompare}. WD-GLT includes edges not involved in the training data in its loss function. This improves its sparsification capability compared to UGS on the Cora and PubMed datasets. While Fast-GLT gives competitive results, especially on PubMed, it loses considerable accuracy for higher sparsification ratios. On the other hand, Shapley value based GNNShap achieves significantly higher sparsity with minimal loss in accuracy.

The results show us that Shapley value based GNN explanations are better suited for graph sparsification due to the following limitations of GLT methods: (i) they require model training for each pruning percentage, and (ii) they have limited sparsification capability because of the necessity of labeled data. However, explainability approaches can learn the importance of edges for predicted classes, which eliminates the need for labels and enables higher pruning percentages with minimal loss of accuracy. Moreover, once edge scores are computed, there is no need to recompute edge scores for each sparsity level, making finding the ideal sparsity level much more effortless. A downside of using graph explainability scores to prune graphs is that it cannot sparsify model weights. However, starting with a small model and gradually increasing the model size until reaching a good accuracy requires less effort than starting with a large model and finding the ideal sparsity level by training the model multiple times.

\subsection{Ablation Study}
\begin{figure*}[!t]
  \centering
    \includegraphics[width=0.90\textwidth]{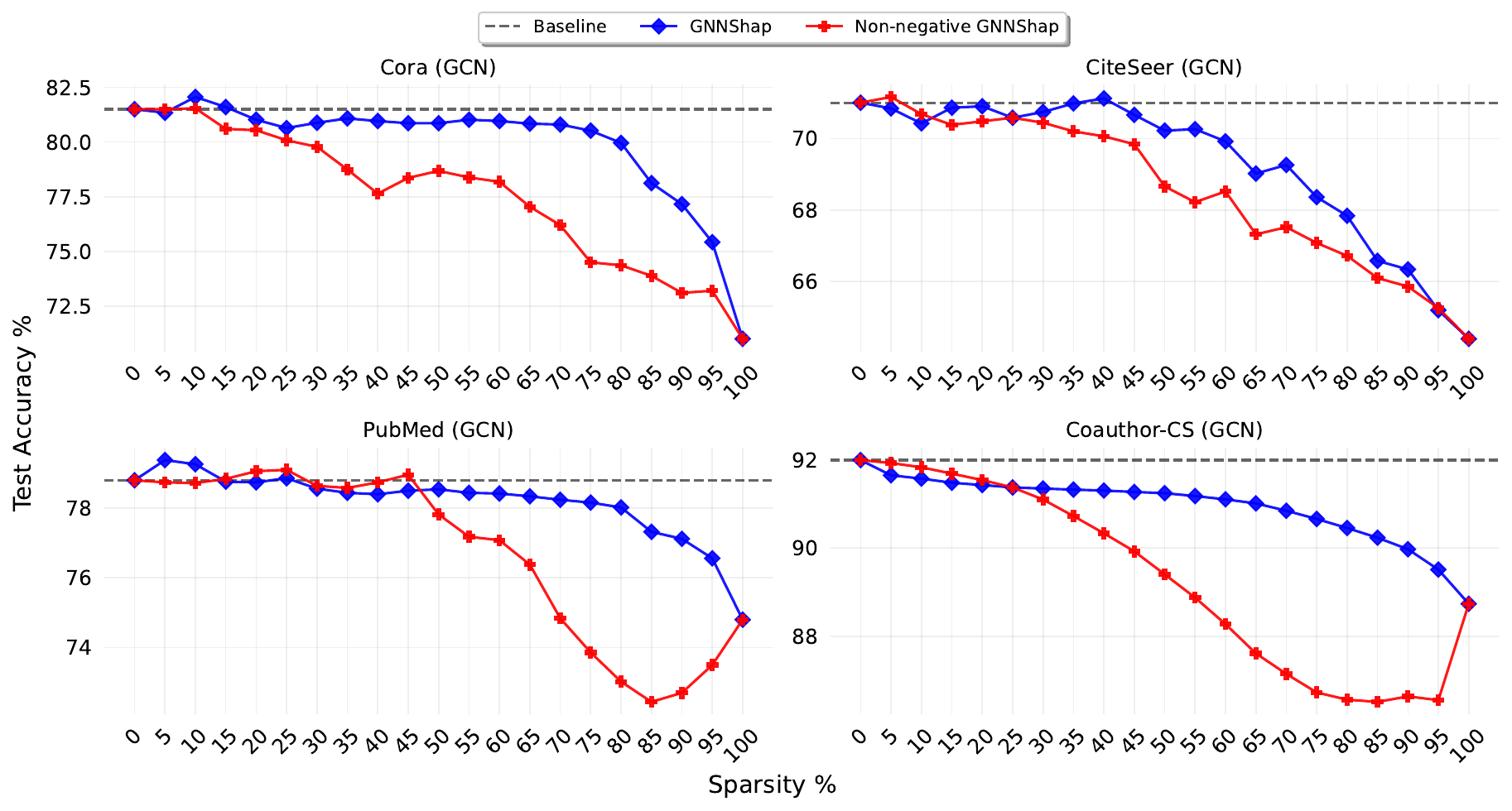}
    \caption{GNNShap test accuracy comparison when non-negative explanation scores are used. Non-negative scores significantly reduce test accuracy.}
  \label{fig:sparse-nonneg}
\end{figure*}

In this section, we investigate the effect of positive and negative attribution scores compared to non-negative explanation scores. For the non-negative GNNShap, we take the absolute value of GNNShap's scores and then compare the test accuracies. Figure~\ref{fig:sparse-nonneg} shows a significant decrease in GNNShap's pruning effectiveness when non-negative scores are utilized for pruning. Considering negatively attributed edges as important (and thus not pruning them) introduces noise to the sparsified graph and reduces the pruning capability of GNNShap.

\section{Conclusion}
In this work, we have investigated the usability of Shapley values for graph sparsification. Shapley values provide both positive and negative explanation scores. This scoring mechanism enables more effective graph sparsification, thereby enhancing the efficiency and scalability of GNNs without compromising accuracy.

Our extensive evaluation demonstrates that Shapley value based sparsification achieves superior accuracy for more significant sparsification percentages, outperforming existing methods on three out of four datasets and across two models. Additionally, Shapley value based sparsification shows better sparsification ratios than graph lottery ticket approaches, highlighting its efficiency in reducing graph complexity.

However, a limitation of using explanation scores in sparsification is that if the underlying model does not perform well, the explanations generated can be misleading, as they are based on incorrect predictions. This limitation affects the applicability of Shapley values, as the reliability of explanations depends on the accuracy of the model.

In conclusion, Shapley value based graph sparsification successfully identifies important edges and provides more effective sparsification while maintaining the accuracy of GNNs. Future work can be designing a more effective aggregation scheme to combine local explanation scores with global importance scores. The improved aggregation scheme can enhance the reliability and applicability of Shapley values, resulting in even higher sparsities without compromising accuracy.

\section{Acknowledgements}
This research is partially supported by the Applied Mathematics Program of the DOE Office of Advanced Scientific Computing Research under contracts numbered DE-SC0022098 and DE-SC0023349 and by NSF grants CCF-2316234 and OAC-2339607.

\bibliographystyle{cas-model2-names}
\bibliography{references}

\appendix
\section{More Aggregation Methods}
\label{sec:sparse-appendix-more-aggr}

\begin{figure*}[htb]
  \centering
    \includegraphics[width=\textwidth]{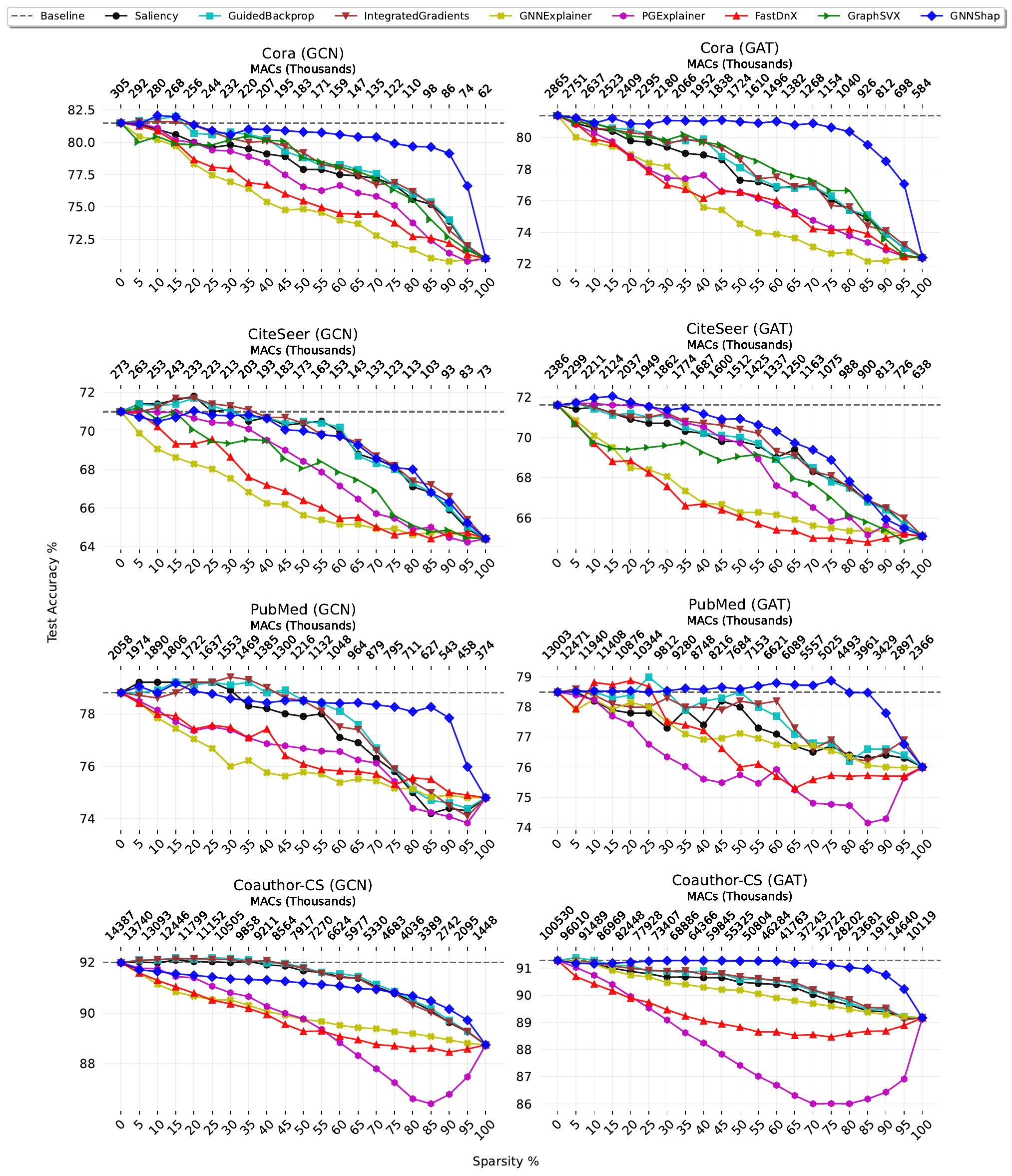}
    \caption{Test accuracies when edges sparsified using \textbf{sum} aggregated explanation scores. GNNShap gives competitive or even better accuracies for high sparsification percentages.}
  \label{fig:sparse-sumres}
\end{figure*}

\begin{figure*}[htb]
  \centering
    \includegraphics[width=\textwidth]{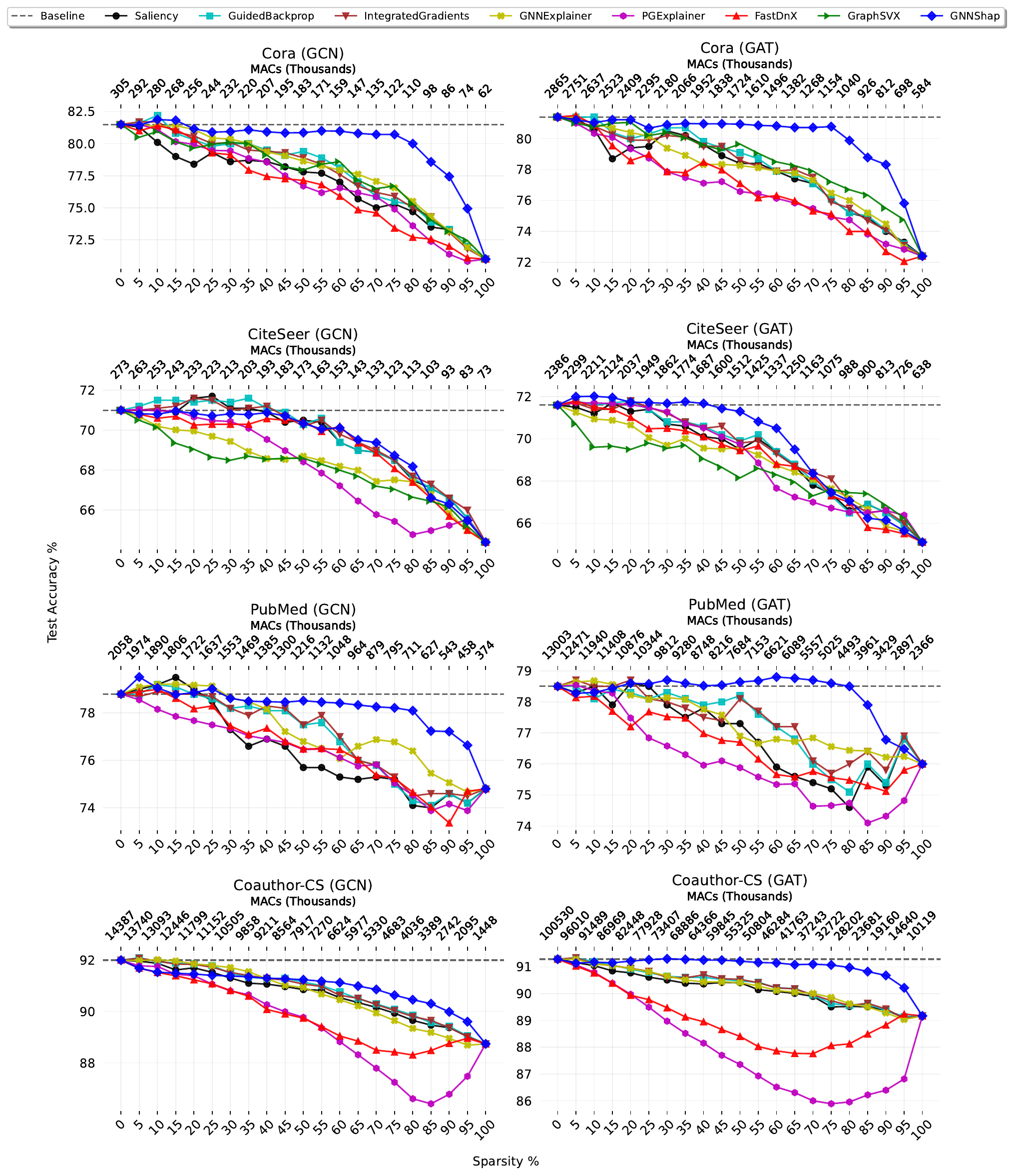}
    \caption{Test accuracies when edges sparsified using \textbf{weighted mean} aggregated explanation scores. GNNShap gives competitive or even better accuracies for high sparsification percentages.}
  \label{fig:sparse-wmeanres}
\end{figure*}

We provide two alternative aggregation results: \textbf{sum} and \textbf{weighted mean}. While sum uses the sum of scores as a global mask, the weighted mean uses model prediction as weights and applies the weighted mean. Figure \ref{fig:sparse-sumres} and \ref{fig:sparse-wmeanres} show these aggregation results. However, we don't see a significant difference compared to mean aggregation.

\clearpage
\end{document}